# ABOUT AN AUTOMATING ANNOTATION METHOD FOR ROBOT MARKERS

Wataru Uemura and Takeru Nagashima

Faculty of Advanced Science and Technology, Ryukoku University, Shiga, Japan

## ABSTRACT

*Factory automation has become increasingly important due to labor shortages, leading to the introduction of autonomous mobile robots for tasks such as material transportation. Markers are commonly used for robot self-localization and object identification. In the RoboCup Logistics League (RCLL), ArUco markers are employed both for robot localization and for identifying processing modules. Conventional recognition relies on OpenCV-based image processing, which detects black-and-white marker patterns. However, these methods often fail under noise, motion blur, defocus, or varying illumination conditions.*

*Deep-learning-based recognition offers improved robustness under such conditions, but requires large amounts of annotated data. Annotation must typically be done manually, as the type and position of objects cannot be detected automatically, making dataset preparation a major bottleneck. In contrast, ArUco markers include built-in recognition modules that provide both ID and positional information, enabling automatic annotation.*

*This paper proposes an automated annotation method for training deep-learning models on ArUco marker images. By leveraging marker detection results obtained from the ArUco module, the proposed approach eliminates the need for manual labeling. A YOLO-based model is trained using the automatically annotated dataset, and its performance is evaluated under various conditions. Experimental results demonstrate that the proposed method improves recognition performance compared with conventional image-processing techniques, particularly for images affected by blur or defocus. Automatic annotation also reduces human effort and ensures consistent labeling quality. Future work will investigate the relationship between confidence thresholds and recognition performance.*

## 1. INTRODUCTION

In recent years, factory automation has become increasingly important due to labor shortages. Markers are often used for self-localization and object identification of these robots. In the RoboCup Logistics League (RCLL) [1, 2], a robotics competition focusing on the automation of logistics in factories and warehouses, ArUco markers [3] are used for robot self-localization and for identifying processing modules. By capturing images of these markers and applying image processing techniques such as those provided by OpenCV [4], the ID of each marker can be recognized based on its black-and-white pattern.

However, OpenCV-based image processing methods may fail to perform accurately in situations with significant noise, such as motion blur or defocus that occur while the robot is moving [5]. To address this issue, this paper focuses on image recognition using deep learning [6-10]. Deep learning enables the identification of unknown images that have correlations with previously learned images. By training the model on images captured under varying conditions such as brightness and viewing angle, it becomes possible to recognize images that are difficult to identify using conventional OpenCV-based methods.

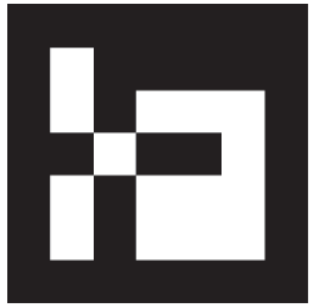

Figure 1: Original ArUco marker with 5x5 cells which ID is 101.

In deep learning, annotation is required to specify the type and position of target objects in training and evaluation images. Since object positions cannot usually be detected automatically, annotation must be performed manually. However, because a large amount of image data is typically required for training, this manual process imposes a significant burden on human operators.

Since markers such as ArUco have built-in recognition modules, it is possible to obtain both the position and ID information of target objects, making automatic annotation feasible. Leveraging this property, this paper proposes a method to automate the annotation process during the training of marker images. This approach significantly reduces the manual effort required for annotation and ensures consistent labeling quality.

Related work has explored both traditional image-processing techniques, such as thresholding and contour extraction, for recognizing fiducial markers including ArUco and AprilTag, and more recent deep-learning-based approaches. Although traditional methods are computationally efficient, their performance degrades under motion blur, defocus, and illumination variations. Deep-learning-based methods improve robustness under such conditions, but they require large amounts of manually annotated training data. The proposed method addresses this issue by enabling automated annotation specifically for ArUco markers.

The remainder of this paper is organized as follows. Section 2 describes the ArUco marker and its detection process. Section 3 explains the CNN-based learning model, including YOLO [11] and its evaluation. Section 4 presents the proposed automated annotation method. Section 5 discusses the experimental setup, results, and analysis. Finally, Section 6 concludes the paper.

## 2. MARKER RECOGNITION

### 2.1. ArUco Marker

In this paper, ArUco markers [3] are detected and identified using the ArUco module of OpenCV [4]. ArUco markers are primarily used for object identification and self-localization of autonomous mobile robots. An ArUco marker is a square-shaped marker, as shown in Figure 1, with its interior composed of black and white square cells and the outer border in black. This structure is similar to that of a QR code [12-13]. The high contrast between black and white cells provides excellent recognition performance in image processing. Compared to QR codes, ArUco markers have a simpler structure, which results in lower information capacity but faster processing speed.

Each ArUco marker is assigned a unique ID based on its internal cell pattern, and this ID is used for marker identification. During detection, the black border of the marker is extracted, and during identification, the internal cell pattern is analyzed to determine the ID. Furthermore, by obtaining information on the marker's size and orientation within the image, it is possible to calculate the relative distance and angle between the camera and the marker.

## 2.2. Marker Detection Process

The identification of ArUco markers can be broadly divided into two steps: contour extraction and analysis of the interior pattern. In the contour extraction step, the captured image is binarized, and black, square-like contours are extracted from the image. However, the extracted contours are not always perfect squares; depending on the viewing angle, they may appear distorted, such as trapezoidal shapes. To address this, a perspective transformation is applied to convert the marker contours back to their original square shape. This allows markers captured from different positions and angles to be processed using a common reference.

Next, the interior of the square-transformed contour is binarized using Otsu's thresholding method [14-15]. Otsu's method automatically determines the threshold by analyzing the histogram of the image and maximizing the between-class variance. Since the brightness of markers can vary depending on the lighting conditions during image capture, applying a fixed threshold to all images is undesirable. After these preprocessing steps, the analysis of the interior pattern is performed.

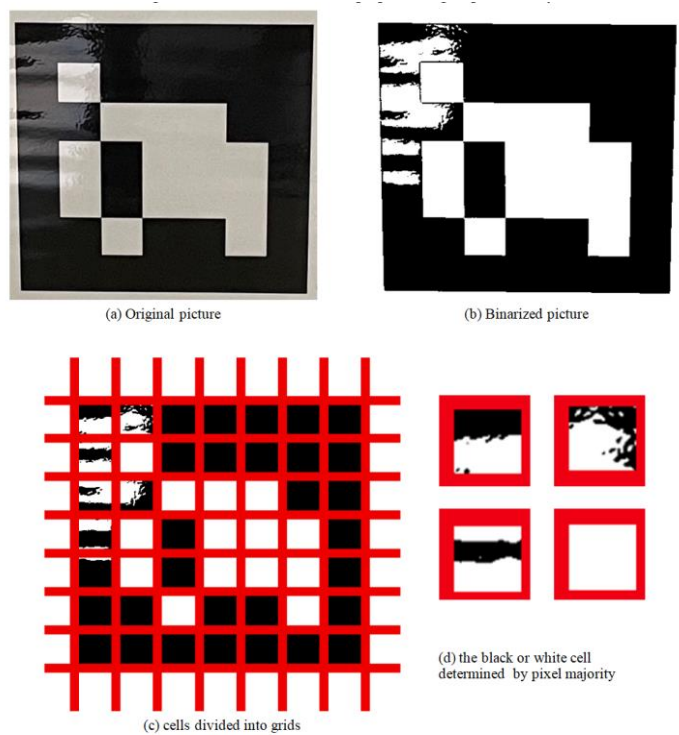

(a) Original picture  (b) Binarized picture

(c) cells divided into grids  (d) the black or white cell determined by pixel majority

Figure 2: How to decide the cells' color.

In the analysis of the interior pattern, the bit pattern inside the contour is examined to verify whether it corresponds to an ArUco marker. As shown in Figure 2, the interior of the contour is divided into cells according to the marker's size. For each cell, the number of black and white pixels is counted, and the cell color is determined by majority voting. The resulting black-and-white pattern is then converted into a bit sequence, which is compared with the registered marker IDs to identify the marker. ArUco markers have error correction capabilities, and the number of correctable bits depends on the number of cells in the marker and the total number of registered markers.

### 2.3. Limitations of Marker Detection

ArUco markers may fail to be identified due to various factors. For example, when a robot captures images of markers while moving, motion blur or defocus can cause identification failures. Other factors include variations in lighting conditions and noise such as dirt or occlusions on the marker surface.

## 3. LEARNING MODEL: CNN AND YOLO

### 3.1. Overview of Convolutional Neural Network (CNN)

In this paper, a Convolutional Neural Network (CNN) [16] is used to train the recognition of ArUco markers. CNN is a type of deep learning method primarily used for image recognition and object detection. The basic structure of a CNN consists of convolutional layers, pooling layers, and fully connected layers. The convolutional layer is responsible for extracting features from the input image. By applying filters and performing convolution operations while sliding across the entire image, it captures features such as edges, colors, and shapes.

The pooling layer compresses the features extracted by the convolutional layer, reducing computational cost while preserving important information. Max pooling, which is commonly used, selects the maximum value within a certain region, providing robustness against shifts in the image.

The fully connected layer classifies objects based on the features extracted by the convolutional and pooling layers. Here, a class refers to the type of object that the deep learning model is designed to recognize.

The structure of deep learning models may vary, leading to differences in object detection and recognition mechanisms. However, the basic structure involves feature extraction using convolutional layers followed by classification based on these features. Depending on the model, some can only recognize the type of object, while others can estimate both the type and the position of the object.

In supervised deep learning, annotation is essential because the model learns to map input images to desired outputs based on ground-truth labels. Without specifying the object class and its location in each training image, the network cannot compute the loss function or update its parameters appropriately. Therefore, annotation is a fundamental requirement for training object detection models such as CNN-based frameworks.

Machine Learning and Applications: An International Journal (MLAIJ) Vol.12, No.4, December 2025

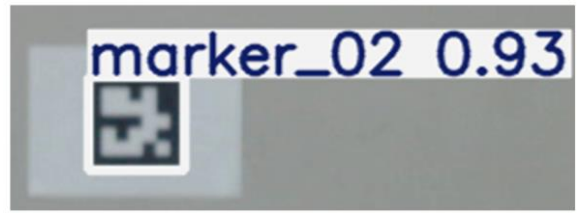

Figure 3: Detection example which a confidence score is 0.93.

### 3.2. You Only Look Once (YOLO) Framework

In this paper, YOLO [8] is used for training marker recognition. YOLO is a type of deep learning model commonly used for image recognition and object detection. For each detected object, YOLO calculates a confidence score. By setting a confidence threshold, an object is considered detected if its score exceeds the threshold. A higher threshold reduces false detections but may decrease the detection rate, whereas a lower threshold increases the detection rate but may result in more false detections.

CNN-based models, including YOLO, were chosen because they are highly effective in extracting spatial features from images and are widely used in real-time object detection tasks. YOLO, in particular, offers fast inference speed and robustness to image distortions such as motion blur and illumination changes, which are common in mobile robot environments. These characteristics make CNN-based detection well suited for recognizing ArUco markers under challenging conditions.

### 3.3. Learning and Evaluation

Here, the learning procedure using CNN is described. Training requires both training and validation images. Training images are used to teach the model features of the target objects, while validation images are used to evaluate the model's performance during training.

These images are captured under varying conditions such as different lighting and viewing angles. Data augmentation techniques can be applied programmatically to create a more generalizable model.

During training, annotation is required for both training and validation images. Annotation involves specifying the type and position of each target object in the image. The learning process uses each image along with its corresponding annotation information.

Test images are used to evaluate the performance of the trained model. Since it is necessary to assess whether the model is effective on unseen data, images that were not used for training or validation are employed for testing. For models capable of estimating both object type and position, a bounding box indicating the region containing the object is generated. Within this region, the model outputs the object type and a confidence score for the detection. In the example shown in Figure 3, a marker with ID 02 is identified within a white bounding box, and the corresponding confidence score of the detection is 0.93.

### 3.4. Limitations of Manual Annotation

In constructing image recognition models using deep learning, annotation is required to specify the type and position of target objects in each image. Since annotation is typically performed manually, it imposes a significant burden on human operators, especially when training requires a large amount of image data.

## 4. PROPOSED AUTOMATED ANNOTATION METHOD

In this paper, we propose a method to automate annotation during marker training. Annotation requires information specifying the type of object and its position in the image. When targeting markers, the recognition module of the marker can provide both the position and the ID, which indicates the type of marker. By using this information, annotation can be automated, enabling more efficient training of markers. This approach reduces the manual effort required when annotations were previously performed by hand.

## 5. EXPERIMENTS AND RESULTS

### 5.1. Experimental Condition

To evaluate the recognition performance of markers trained using the proposed method, we compare the detection and identification rates of images captured during robot movement using conventional OpenCV-based image processing and the proposed automated-annotation-based method. We also investigate the relationship between confidence threshold and recognition performance. In the experiments, the recognition rate is defined as the proportion of markers in the test images that are correctly detected and identified. The misidentification rate is defined as the proportion of markers that are detected but assigned an incorrect ID, or of non-marker objects that are incorrectly detected as markers. Additionally, we assess the potential of the proposed automated-annotation-based method to recognize images that cannot be correctly identified using conventional image processing methods.

28 types of ArUco markers used in the RCLL are employed in the experiments. A Logitech HD Pro Webcam C920 is used as the camera. YOLO, a relatively fast deep learning model, is used for training. Training, validation, and test images are captured with varying amounts of motion blur, as well as different angles and distances between the camera and the markers. For each marker type, approximately 130 images are used for training, 25 for validation, and 200 for testing.

In the experiments, the confidence threshold is varied from 0.3 to 0.8 in increments of 0.1, and the recognition and misidentification rates are measured at each threshold.

Table 1: Parameters and Variation Ranges for Data Augmentation

| Parameter | Our program | YOLO |
|---|---:|---:|
| Brightness | ±100.0 % | ±40.0 % |
| Contrast | ±50.0 % | (none) |
| Hue | ±2.8 % | ±1.5 % |
| Saturation | ±10.0 % | ±70.0 % |
| Scale | +50.0 % | ±50.0 % |
| Location | Random | (none) |
| Translation (x, y) | (none) | ±10 % |
| Mosaic | (none) | 100 % |

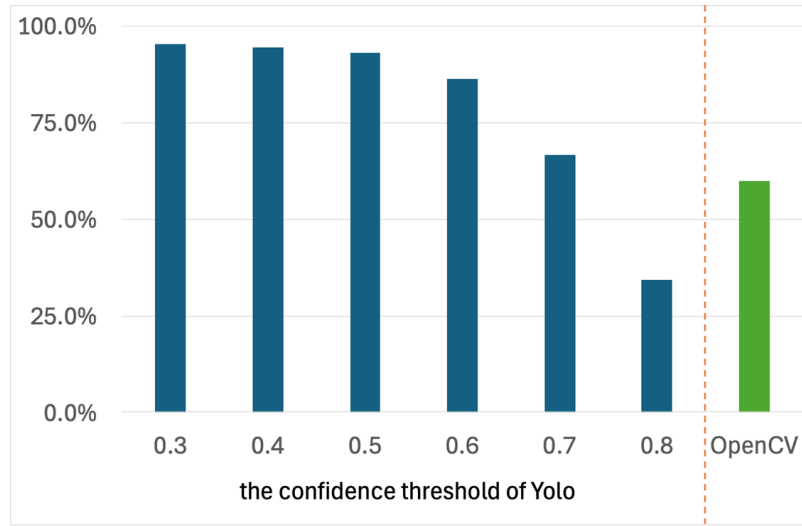

Figure 4: Recognition rates for the conventional OpenCV-based image-processing method and the proposed automated-annotation-based method

### 5.2. Results and Discussion

The recognition rates using the conventional OpenCV-based image processing method and the proposed automated-annotation-based method are shown in Figure 4, and the misidentification rates are shown in Figure 5. It was confirmed that increasing the confidence threshold decreases the recognition rate but reduces misidentifications, whereas decreasing the threshold increases the recognition rate but also increases the misidentification rate.

The proposed method successfully reduced the time cost of annotation through automation. In this experiment, when the confidence threshold was set between 0.4 and 0.7, the recognition rate improved and the misidentification rate decreased compared to the conventional OpenCV-based image processing method. This suggests that the learning-based approach can recognize images that were previously difficult to identify using traditional image processing methods. Most of the images that could not be identified by OpenCV exhibited motion blur or defocus, indicating that learning can potentially mitigate the effects of these factors.

One possible reason for the improved performance under motion blur is that, during CNN processing, the feature information is compressed in a way that absorbs the effects of blur. Since CNNs perform compression of extracted features, it is possible to obtain similar feature representations even when some blur occurs.

Although the proposed method improves robustness against blur and defocus, its performance still depends on the variety of the training dataset. In addition, extremely severe motion blur or very small marker sizes remain challenging for reliable recognition.

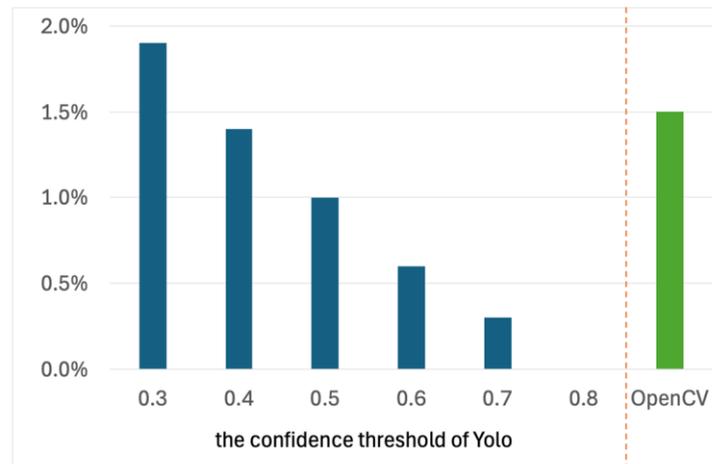

Figure 5: Misidentification rates for the conventional OpenCV-based image-processing method and the proposed automated-annotation-based method

## 6. CONCLUSIONS

In this paper, we proposed a method to automate annotation during marker training, reducing the manual effort required for annotation. The experiments confirmed the relationship between the confidence threshold and recognition performance. The results demonstrate the superiority of the proposed method compared with conventional OpenCV-based processing, particularly under blurred or noisy conditions. The automated annotation approach also reduces dataset preparation time substantially. Furthermore, the results suggest the potential to recognize images affected by motion blur or defocus, which are difficult to handle using conventional image processing methods. Future work includes expanding the method to other types of fiducial markers, improving robustness under extreme blur, and investigating adaptive threshold optimization for confidence-based recognition.

## CONFLICTS OF INTEREST

The authors declare no conflict of interest.

## ACKNOWLEDGEMENTS

This research was partially supported by the Research Program of Ryukoku University.

## REFERENCES

[1] Niemueller, T.; Lakemeyer, G.; and Ferrein, A. (2015). "The RoboCup logistics league as a benchmark for planning in robotics", Planning and Robotics (PlanRob-15), 63.

[2] Dissanayaka, S.; Ferrein, A.; Hofmann, T.; Nakajima, K.; Sanz-Lopez, M.; Savage, J.; Swoboda, D.; Tschesche, M.; Uemura, W.; Viehmann, T.; and Yasuda, S., (2025) "From production logistics to smart manufacturing: The vision for a new RoboCup industrial league", In RoboCup 2025: Robot World Cup XXVIII, 2025 (to appear).

[3] Garrido-Jurado, S.; Muñoz-Salinas, R.; Madrid-Cuevas, F.J.; and Marín-Jiménez, M.J., (2014) "Automatic generation and detection of highly reliable fiducial markers under occlusion", Pattern Recognition, Vol. 47, No. 6, pp. 2280–2292.

[4] Bradski, G., (2000) "The OpenCV Library", Dr. Dobb's Journal of Software Tools, Vol.25, No.11, pp. 122-125.

[5] Nakajima, K.; Nagashima, T.; Komori, Y.; Yasuda, S.; Tanabe, R.; and Uemura, W., (2024) "About an Error Correcting Method of ArUco Markers Considering Row Data for a Mobile Robot", 2024 IEEE 13th Global Conference on Consumer Electronics (GCCE), pp. 273-275.

[6] LeCun, Y.; Bengio, Y.; and Hinton, G., (2015). "Deep learning". Nature. Vol. 521, pp. 436-444.

[7] Bengio, Y. (2016) "Learning Deep Architectures for AI", Foundations and Trends in Machine Learning, Vol. 2, No. 1, pp. 1-127.

[8] Mohammed A.; and Khalid Z., (2024) "Evaluating the Impact of Convolutional Neural Network Layer Depth on the Enhancement of Inertial Navigation System Solutions", International Journal of Computer Networks & Communications (IJCNC), Vol. 16, No. 5, pp. 59-76.

[9] Agarwal, P.; Bhasin, A.; Keshwani, R.; Verma, A.; and Kaushik, S., (2016), "Machine Learning Toolbox", Machine Learning and Applications: An International Journal (MLAIJ), Vol. 3, No. 3, pp. 25-34.

[10] Mamdouh, M. G.; Alaa, E.; Mahmoud M. E.; and Alaa M. Z., (2023), "Face Mask Detection Model Using Convolutional Neural", Machine Learning and Applications: An International Journal (MLAIJ), Vol. 10, No. 2/3, pp. 33-46.

[11] Redmon, J.; Divvala, S.; Girshick, R.; and Farhadi, A., (2016), "You Only Look Once: Unified, Real-Time Object Detection", 2016 IEEE Conference on Computer Vision and Pattern Recognition (CVPR), pp. 779-788.

[12] Hara, M.; Watabe, M.; Nojiri, T.; Nagaya, T.; and Uchiyama, Y., (1995), "Two-dimensional code", JPH07254037A.

[13] Hara, M.; Watabe, M.; Nojiri, T.; Nagaya, T.; and Uchiyama, Y., (1999), "2D code", JP2938338B2.

[14] Otsu, N., (1979), "A threshold selection method from gray-level histograms", IEEE Transactions on Systems, Man, and Cybernetics, Vol.9, No. 1, pp. 62-66.

[15] Sezgin, M.,; and Sankur, B., (2004), "Survey over image thresholding techniques and quantitative performance evaluation", Journal of Electronic Imaging. Vol. 13, No. 1, pp. 146–165.

[16] Venkatesan, R.; and Li, B., (2017), "Convolutional Neural Networks in Visual Computing", A Concise Guide, CRC Press.

## AUTHORS

**Wataru Uemura** was born in 1977, and received B.E., M.E., and D.E. degrees from Osaka City University in 2000, 2002, and 2005. He is an associate professor of the Faculty of Advanced Science and Technology, Ryukoku University in Shiga, Japan. He is a member of IEEE, RoboCup, and others. He is the chairperson of RoboCup Japanese Regional Committee.

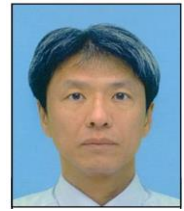

**Takeru Nagashima** was born in 2003 and received B.E. from Ryukoku University in 2025. He is interested in Computer vision and Machine Learning.

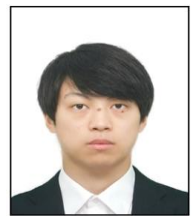